# DISTRIBUTION OF THE SEARCH OF EVOLUTIONARY PRODUCT UNIT NEURAL NETWORKS FOR CLASSIFICATION

Antonio J. Tallón-Ballesteros
*Department of Languages and Computer Systems. University of Seville*
*Numberless Reina Mercedes Avenue. Seville, 41012 Spain*
atallon@us.es

Pedro A. Gutiérrez-Peña, César Hervás-Martínez
*Department of Computer Science and Numerical Analysis. University of Córdoba*
*Campus of Rabanales. Córdoba, 14071 Spain*
zamarck@yahoo.es, chervas@uco.es

**ABSTRACT**

This paper deals with the distributed processing in the search for an optimum classification model using evolutionary product unit neural networks. For this distributed search we used a cluster of computers. Our objective is to obtain a more efficient design than those net architectures which do not use a distributed process and which thus result in simpler designs. In order to get the best classification models we use evolutionary algorithms to train and design neural networks, which require a very time consuming computation. The reasons behind the need for this distribution are various. It is complicated to train this type of nets because of the difficulty entailed in determining their architecture due to the complex error surface. On the other hand, the use of evolutionary algorithms involves running a great number of tests with different seeds and parameters, thus resulting in a high computational cost.

## 1. INTRODUCTION

This research is about the distribution of processing involved in the search for the best product-unit neural network (PUNN) models [Durbin, 1990] [Martínez-Estudillo, 2006A], using evolutionary algorithms, EAs. A cluster of computers [Buyya, 1999] will be used to carry out the distribution of this processing.

Many different types of neural network architectures have been used, but the most popular one has been the single-hidden-layer feedforward network. Amongst the numerous approaches that use neural networks in classification problems, we focus our attention on evolutionary artificial neural networks (EANNs). EANNs have been a key research area in the past decade providing an improved platform for optimizing network performance and architecture (number of hidden nodes and number of connections) simultaneously. [Miller, 1989] proposed that evolutionary computation was a very good candidate for searching the space of architectures because the fitness function associated with that space is complex, noisy, non-differentiable, multi-modal and deceptive. Since then, many evolutionary programming methods have been developed to evolve artificial neural networks, for instance [Yao, 1997] and [García-Pedrajas, 2002].

An objective of this paper is to design a neural network architecture that will be suitable for classification; this involves achieving a proper balance between the explorative/exploitative activities performed by a learning EA. The design of a search for good EANN models is usually based on trial and error (that is, exploring the number of nodes in the hidden layer and exploring the different parameters of the EA); the strategy follows a blind search using only a small group of possible configurations. This implies a high computational cost in order to solve a medium-sized problem, as well as the use of an EA in the search process, which makes the task practically overwhelming for even a well-equipped computer. In our case, we distribute certain network or EA parameters throughout the cluster, so that we can carry out several experiments in about the time we previously would have needed for only one.

Different reasons make this distribution necessary: i) the training of this type of nets is complicated by the difficulty entailed in determining the architecture of the models (number of hidden layers, number of connections), due to the complexity of the error surface; ii) the use of an EA (stochastic search) demands a high number of runs using different seeds and parameters, giving rise to a high computational cost. The fundamental goal of this work is twofold: to improve efficiency, through the distribution of different values for the main parameters in the EA in different nodes of the cluster, and also to improve the speedup in the computation in different runs searching for models.

The rest of this paper is organized as follows. In section 2 we describe the proposed methodology. Section 3 continues with the distributed computation of the design of neural networks and the analysis of the results obtained. Finally, in the last section we present concluding remarks.

## 2. DESCRIPTION

The proposed methodology consists of the use of an EA as a tool for learning the architecture and weights of a PUNN ([Martínez-Estudillo, 2006A], [Martínez-Estudillo, 2006B]). This class of multiplicative neural networks comprises such types as sigma-pi networks and product unit networks. Some advantages of PUNNs are increased information capacity and the ability to form higher-order combinations of the inputs [Durbin, 1990]. Besides that, it is possible to obtain the upper bounds of the VC dimension of product-unit neural networks similar to those obtained for sigmoidal neural networks [Schmitt, 2001]. Finally, it is a straightforward consequence of the Stone-Weierstrass Theorem to prove that product-unit neural networks are universal approximators [Martínez-Estudillo, 2006A]. Despite these advantages, product-unit based networks have a major drawback. Networks based on Product Units (PUs) have more local minima and more probability of becoming trapped in them [Ismail, 2000]. The main reason for this difficulty is that small variations in the exponents can cause large changes in the total error surface. Several efforts have been made to carry out learning methods for PUs. [Janson, 1993] developed a genetic algorithm for evolving the weights of a network based on PUs with a predefined architecture. The major problem of this kind of algorithm is how to obtain the optimal architecture before hand [Ismail, 2000].

The disadvantage of using EA in the training of PUNN is that the processing time could be too great with respect to the dimension of the characteristics space and to the number of classes considered in a concrete problem. So, a system has been created to allow the configuration parameter distribution throughout the different nodes in the cluster of computers.

### 2.1 Evolutionary Algorithm

We use an EA to design the structure and learn the weights of PUNNs. The search begins with a random initial population, and, for each iteration, the population is updated using a population-update algorithm. The population is subjected to the operations of replication and mutation. Crossover is not used due to its potential disadvantages in evolving artificial networks. When using the algorithm in classification problems, we will only consider the construction of the aptitude function and the changes involved in mutation operators.

A standard softmax activation function is used for each node of the output layer given by:

$$g_j(\mathbf{x}) = \frac{e^{f_j(\mathbf{x})}}{\sum_{i=1}^{J} e^{f_i(\mathbf{x})}} \qquad (1)$$

where $l$ is the number of classes, $f_j(\mathbf{x})$ is the output of the node j for the pattern $\mathbf{x}$ and $g_j(\mathbf{x})$ is the probability of this pattern belongs to the class *j (C$_j$)*. Taking this into account, the function of cross-entropy error is used to evaluate individuals for a network R which is reflected in the following expression:

$$l(R) = -\frac{1}{n} \cdot \sum_{i=1}^{n} \sum_{j=1}^{l} (y_i^j \ln(g_j(\mathbf{x}_i))) \qquad (2)$$

and substituting $g_j$ defined in (1),

$$l(R) = \frac{1}{n} \cdot \sum_{i=1}^{n} \left( -\sum_{j=1}^{l} y_i^j f_j(\mathbf{x}_i) + \ln(\sum_{j=1}^{l} e^{f_j(\mathbf{x})}) \right) \qquad (3)$$

where $y_i^j$ is the target value for the class *j* in pattern *i* ( $y_i^j = 1$ if $\mathbf{x} \in C_j$ and $y_i^j = 0$ otherwise), $f_j(\mathbf{x})$ the output value of the neural network for the output neuron *j* with pattern *i* and *n* the number of patterns.

Since we want the EA to minimise the chosen error function, we use a fitness function in the form $A(R) = (1 + l(R))^{-1}$. Parametric mutation consists of a simulated annealing algorithm. The severity of a mutation to an individual $R$ is dictated by the temperature $T(R)$, given by $T(R) = 1 - A(R)$, $0 \le T(R) < 1$. Parametric mutation is accomplished for each coefficient $w_{ji}$, $\beta_j^l$ of the model with Gaussian noise, where the variance depends on the temperature: $w_{ji}(t+1) = w_{ji}(t) + \xi_1(t)$ and $\beta_j^l(t+1) = \beta_j^l(t) + \xi_2(t)$ where $\xi_k(t) \in N(0, \alpha_k T(R))$, $k = 1, 2$, represents a one-dimensional normally distributed random variable with mean 0 and standard deviation $\alpha_k(t) \cdot T(R)$. It should be pointed out that the modification of the exponents $w_{ji}$ is different from the modification of the coefficients $\beta_j^l$, therefore $\alpha_1 << \alpha_2$. Structural mutation implies a modification in the function structure and allows the exploration of different regions in the search space while helping to keep the diversity of the population. There are four different structural mutations similar to the ones in the GNARL model [Angeline, 1994]: node addition, node deletion, connection addition and connection deletion. All the above mutations are made sequentially in the given order, with probability $T(R)$, in the same generation on the same network. If the probability does not select any mutation, one of the mutations is chosen at random and applied to the network. For more details about the EA see ([Martínez-Estudillo, 2006A], [Martínez-Estudillo, 2006B]).

In our case, since what we have is a classification problem, the evolution process must be short. That is why an evolution mechanism must be selected for parameters α$_1$ and α$_2$ that converges toward optimum values more quickly. We use the 1/5 success rule of Rechenberg [Rechenberg, 1973].

## 3. IMPLEMENTED DISTRIBUTED MODELS

### 3.1 Experimental design distribution

In this first case, some parameters of the EA are distributed throughout all the nodes in the cluster. These are the number of neurons in the hidden layer (neu), the maximum number of generations (gen) and optionally also the value of the output-variance (α$_2$). We will consider a base configuration which will be modified by each of the processing nodes. Therefore, each one will have a specific mission and it will always tune the same parameters. The changes have been defined in a relative way, and so these will depend on the base configuration. Once the modifications have been made, each of the nodes will run the experiments with the new configuration. There are 8 nodes, so in the case of distributing 3 parameters, each one of them will take 2 different values and, in the case of 2 parameters, one will have 4 different values and the other one will have 2 values. For this distribution of the processing, we have used the following datasets: Balance, Cancer, Pima, Hypothyroid and Waveform, all of them from the UCI repository

(http://www.ics.uci.edu/~mlearn/MLRepository.html). These databases are summarized in the following table:

Table 1. Summary of the databases used for the experimental design distribution

| Dataset | Total Patterns | Train Patterns | Test Patterns | Input variables | Classes |
|---|---|---|---|---|---|
| Balance | 625 | 469 | 156 | 4 | 3 |
| Cancer | 699 | 525 | 174 | 9 | 2 |
| Pima | 768 | 576 | 192 | 8 | 2 |
| Hypothyroid | 3772 | 2829 | 943 | 29 | 4 |
| Waveform | 5000 | 3750 | 1250 | 40 | 3 |

### 3.1.1 Validation technique and parameters used

We have considered a cross validation experimental design called hold-out [Prechelt, 1994], that consists of splitting the data in two sets: train and test. In our case, the size of the train set is 3n/4 and n/4 for the test set, where n is the number of patterns in the problem. The results of the experimentation in this proposal have been obtained with the following common parameters / features of configuration.

Table 2.- Common configuration parameters / features of the experimental design distribution

| Parameter / Feature | Value / Type |
|---|---|
| Population size | 1000 |
| Range for the weights between input layer and hidden layer | [-5, 5] |
| % of the individuals over which parametrical and structural mutations are applied | 10 % - 90 % |
| Initial values of $\alpha_1$ and $\alpha_2$ | 0.5 and 1 respectively |
| Normalization of the input data | [1, 2] |
| Number of nodes in node addition and node deletion operators | [1, 2] |
| Percentages of links to be mutated in the input and the output layer | 30% and 5% respectively |
| Number of runs | 30 |

The base configuration values depend on the database and number of parameters to distribute. There will be different configurations depending on weather 2/3 parameters are distributed.

Table 3. Configurations of the experimental design distribution with 3 parameters

| Configuration \ Parameters | 1 | 2 | 3 | 4 | 5 | 6 | 7 | 8 |
|---|---|---|---|---|---|---|---|---|
| Neurons | neu | neu + 1 | neu | neu + 1 | neu | neu + 1 | neu | neu + 1 |
| Max. Number of Generations | gen | gen | gen | gen | 0.8 * gen | 0.8 * gen | 0.8 * gen | 0.8 * gen |
| Output-variance ($\alpha_2$) | $\alpha_2$ | $\alpha_2$ | 1.5 * $\alpha_2$ | 1.5 * $\alpha_2$ | $\alpha_2$ | $\alpha_2$ | 1.5 * $\alpha_2$ | 1.5 * $\alpha_2$ |

Table 4. Configurations of the experimental design distribution with 2 parameters

| Configuration \ Parameters | 1* | 2* | 3* | 4* | 5* | 6* | 7* | 8* |
|---|---|---|---|---|---|---|---|---|
| Neurons | neu | neu + 1 | neu + 2 | neu + 3 | neu | neu + 1 | neu + 2 | neu + 3 |
| Max. Number of Generations | gen | gen | gen | gen | 0.8 * gen | 0.8 * gen | 0.8 * gen | 0.8 * gen |

Next, the initial values of the parameters to distribute are shown in Table 5. In Hypothyroid and Waveform, the value of $\alpha_2$ will not be distributed.

Table 5. Values of the base configuration of the experimental design distribution

| Database \ Parameters | Balance | Cancer | Pima | Hypothyroid | Waveform |
|---|---|---|---|---|---|
| Neurons | 5 | 2 | 3 | 3 | 3 |
| Max. Number of Generations | 150 | 100 | 120 | 500 | 500 |
| Output-variance ($\alpha_2$) | 1 | 1 | 1 | - | - |

### 3.1.2 Results

In this section we present the results obtained in the validation set of each of the groups described with 8 configurations. We have called base configuration to the first one of each group. Amongst some of the configurations, the topology (number of neurons in the hidden layer) and the parameters (variance -$\alpha_2$-, maximum number of generations) are changed, and amongst others only the topology is changed. The best configuration is seen here in bold. We will show the mean values of the generalization CCR of the best model obtained, its standard deviation, the best and the worst value. The topologies will be shown with this format: *Number of inputs: Hidden layer node number: Output layer node number*.

➢ Medium size datasets (Balance, Pima and Cancer).- The following results have been obtained:

Table 6. Results of the experimental design distribution with medium size datasets

|  |  | Generalization CCR | | | | | | | |
|---|---|---|---|---|---|---|---|---|---|
| **Balance** | Config. (Topology) | 1 (4: 5: 2) | 2 (4: 6: 2) | 3 (4: 5: 2) | **4 (4: 6: 2)** | 5 (4: 5: 2) | 6 (4: 6: 2) | 7 (4: 5: 2) | 8 (4: 6: 2) |
|  | Mean | 95.2991 | 95.1495 | 95.0427 | **95.6196** | 94.5512 | 94.7008 | 94.4658 | 94.5512 |
|  | Std. Dev. | 1.4707 | 0.9770 | 1.4077 | **1.1551** | 1.7635 | 1.6315 | 1.3931 | 1.3228 |
|  | Best | 98.0769 | 96.7948 | 97.4358 | **98.0769** | 98.7179 | 98.0769 | 96.7948 | 96.7948 |
|  | Worst | 92.9487 | 93.5897 | 92.3076 | **93.5897** | 90.3846 | 91.6666 | 92.3076 | 91.6666 |
| **Cancer** | Config. (Topology) | 1 (9: 2: 1) | 2 (9: 3: 1) | 3 (9: 2: 1) | 4 (9: 3: 1) | 5 (9: 2: 1) | 6 (9: 3: 1) | 7 (9: 2: 1) | **8 (9: 3: 1)** |
|  | Mean | 98.4865 | 98.9655 | 98.5057 | 98.716 | 98.5632 | 98.8122 | 98.3716 | **98.9655** |
|  | Std. Dev. | 0.6127 | 0.3818 | 0.4913 | 0.5977 | 0.5596 | 0.5427 | 0.6593 | **0.3818** |
|  | Best | 99.4252 | 99.4252 | 99.4252 | 100.0000 | 99.4252 | 99.4252 | 99.4252 | **99.4252** |
|  | Worst | 97.1264 | 98.2758 | 97.7011 | 97.7811 | 97.7011 | 97.7011 | 97.1264 | **98.2758** |
| **Pima** | Config. (Topology) | 1 (8: 3: 1) | **2 (8: 4: 1)** | 3 (8: 3: 1) | 4 (8: 4: 1) | 5 (8: 3: 1) | 6 (8: 4: 1) | 7 (8: 3: 1) | 8 (8: 4: 1) |
|  | Mean | 77.3263 | **78.6111** | 76.9618 | 77.6909 | 77.1527 | 77.2222 | 76.7534 | 77.0312 |
|  | Std. Dev. | 2.3600 | **1.8800** | 1.6678 | 1.7942 | 1.7014 | 2.2388 | 1.9558 | 1.7641 |
|  | Best | 80.2083 | **82.2916** | 80.2083 | 81.2500 | 79.6875 | 81.7708 | 81.7708 | 79.6875 |
|  | Worst | 71.3541 | **73.4375** | 73.9583 | 74.4791 | 72.3958 | 73.4375 | 73.4375 | 72.9166 |

➢ Large datasets (Hypothyroid and Waveform).- The results obtained are:

Table 7. Results of the experimental design distribution with big size datasets

|  |  | Generalization CCR | | | | | | | |
|---|---|---|---|---|---|---|---|---|---|
| **Hypothyroid** | Config. (Topology) | 1* (29:3:4) | 2* (29:4:4) | **3* (29:5:4)** | 4* (29:6:4) | 5* (29:3:4) | 6* (29:4:4) | 7* (29:5:4) | 8* (29:6:4) |
|  | Mean | 95.2677 | 95.3207 | **95.5726** | 95.4334 | 95.1749 | 95.0424 | 95.0424 | 95.2942 |
|  | Std. Dev. | 0.7708 | 0.5764 | **0.4998** | 0.5537 | 0.2890 | 0.2176 | 0.2582 | 0.1596 |
|  | Best | 96.9247 | 96.1823 | **96.0763** | 96.6065 | 95.7582 | 95.2279 | 95.4400 | 95.5461 |
|  | Worst | 94.5917 | 94.5917 | **94.9098** | 94.6977 | 94.9098 | 94.6977 | 94.5917 | 95.1219 |
| **Waveform** | Config. (Topology) | 1* (40:3:2) | 2* (40:4:2) | 3* (40:5:2) | **4* (40:6:2)** | 5* (40:3:2) | 6* (40:4:2) | 7* (40:5:2) | 8* (40:6:2) |
|  | Mean | 82.4400 | 83.5900 | 84.3500 | **85.1100** | 82.4500 | 83.5300 | 83.5700 | 84.4900 |
|  | Std. Dev. | 1.5883 | 1.0364 | 1.5322 | **1.4846** | 0.8357 | 1.1753 | 1.6894 | 0.8107 |
|  | Best | 83.9200 | 84.8800 | 86.8000 | **87.5200** | 83.2800 | 85.2800 | 85.6000 | 85.7600 |
|  | Worst | 79.6800 | 81.9200 | 82.3200 | **82.8800** | 80.6400 | 82.0800 | 81.0400 | 82.9600 |

### 3.1.3 Statistical Analysis

We have performed a statistical analysis to compare the base configuration and the best for each one of the first three above-mentioned databases. In this comparison we have used the CCR and the number of connections obtained in the generalization process.

In Balance, once the algorithm has been run 30 times, using the default parameters (base configuration) and using those that obtained the best results, we present the following considerations: a) using the Kolmogorov-Smirnov (K-S) test, we conclude that the CCR distributions of the validation set and the number of connections in the best network model for each run are distributed with a normal distribution with asintotic signification levels greater than the standard value α = 0.05, b) under normality hypothesis, we did variance equality contrasts (Levene test) and means equality (with equal or different variances depending on the previous Levene test result). These Student's t tests have been done considering that the 30 runs of the base and best algorithm are independent. The results for CCR show that there are no significant differences in variances (Sig = 0.253), nor in mean values (Sig = 0.352). In addition, the results for the number of connections of the best network models reveal that there are no significant differences in the variances (Sig = 0.098), however, there are differences with respect to the connection medium number with α = 0.05 (Sig = 0.000); thus, the models run with the base parameters are less significant and shorter.

The results for Cancer show that if we use the K-S test, we conclude that the CCR distributions of the test set and the number of connections follows a normal distribution with asintotic signification levels greater than the value α = 0.01. On the other hand, under the hypothesis of normal and independent distributions, we deduce that there are significant differences in the variances (Sig = 0.004) and in the CCR mean values (Sig = 0.001) with α = 0.05. Likewise, the results for the number of connections in the best models show that there are no significant differences in the variance, altough there are differences with respect to the average number of connections for α = 0.05 (Sig = 0.000); so the obtained models with best parameters are better regarding to CCR mean. From these results we conclude that it is preferable considering for Cancer as algorithm parameters, the ones of the best configuration.

The results for Pima, again, show through the K-S test that the CCR distribution in the test set and the number of connections follow a normal distribution with a level of asintotic signification greater than α = 0.01. On the other hand, under the hypothesis of normal and independent distributions, we infer that there are no significant differences in the variances (Sig = 0.118); however there are differences in the CCR mean values (Sig = 0.023) with α = 0.05. In the same way, the results for the number of connections with the best network models show that there are no significant differences in the variances (Sig = 0.305); instead there are differences in the mean number of connections for α = 0.05 (Sig = 0.000), so the model obtained with the best parameters are significantly better regarding to greater significant values in the mean CCR, though with respect to the number of connections there is a significant mean number greater that the models with base parameters. From the previous results we conclude that it is preferable for Pima to consider as algorithm parameters, the ones of the best configuration.

### 3.1.4 Comparison to previous results

Although the generalization CCR values seem very good, we compare them to those obtained previously in the standard PUs, that do not apply distribution models. In the Table 8 appears the generalization CCR mean value of the best model obtained with the best configuration using standard PUs and distributed PUs. The best results obtained with the distribution are written in bold.

Table 8. Comparison of the results obtained in the experimental design distribution

| Dataset<br>Methodology | Balance | Cancer | Pima | Hyphotyroid | Waveform |
|---|---|---|---|---|---|
| Standard PUs | 95.38±1.37 | 98.92±0.55 | 77.79±1.47 | 94.25±1.07 | 83.08±0.55 |
| **Distributed PUs** | **95.61±1.15** | **98.96±0.38** | **78.61±1.88** | **95.57±0.49** | **85.11±1.48** |

As we can see, the average of the results obtained in the best distributed PU models overcomes in all cases to the results achieved with standard PUs. In Balance there is a 0.23% of improvement; a variance reduction is produced as well, which indicates a greater homogeneity in the solutions. In Cancer there is a generalization CCR increment of about a 0.04% and the variance decreases slightly. Pima has a more significant improvement (0.82%); however, the variance increases slightly. In Hyphotyroid the generalization CCR increases to 1.32%. In Waveform, the improvements are about 2.03%.
We conclude that the improvement is greater in the bigger databases than in the smaller ones.

## 3.2 Processing distribution

### 3.2.1 Description

In this case we divide the execution number amongst some of the nodes. Our objective is to obtain a measure of the optimal number of nodes considering some of the previous databases. We have used the best configurations of the first three databases in the previous proposal.

### 3.2.2 Performance Analysis

We aim to obtain a time performance measure of the best configurations of the first three databases in the previous proposal. Since a cluster is a parallel or distributed architecture [Buyya, 1999], we will use performance evaluation measures of parallel algorithms ([Kumar, 1994], [Wilkinson, 1999], [Ortega, 2004]), like the speedup and the efficiency. The speedup is defined as the ratio between the execution time in one processor and the parallel time with P equal processors. This definition can be applied to our case, with the consideration that we refer to nodes and not to processors.

$$S = T_1 / T_P$$

In general, $S \leq P$. In practice, the speedup is saturated when the number of processors increases. We will obtain the number of processors from which this saturation takes place. The speedup considers the execution time and it is not always the best measure to evaluate the algorithm throughput, because it varies with the number of processors. It is necessary to normalize the amounts of time to compare the throughput using different nodes. To establish this normalization, we use the efficiency which is defined as the ratio between the speedup and the number of processors. By means of efficiency we can obtain the optimal value regarding to the number of nodes that we must use to carry out experiments. We will consider 32 runs of the best configuration above-mentioned.

In the Table 9 we will show the execution time, the speedup and the efficiency for each of the first three databases.

Table 9. Speedup and efficiency for medium-size datasets

|  | Balance | | | | Cancer | | | | Pima | | | |
|---|---|---|---|---|---|---|---|---|---|---|---|---|
| Number of nodes | 1 | 2 | 4 | 8 | 1 | 2 | 4 | 8 | 1 | 2 | 4 | 8 |
| Time (min.) | 349 | 177 | 88 | 45 | 103 | 53 | 26 | 14 | 215 | 109 | 54 | 29 |
| Speedup ($T_1 / T_p$) | | 1.9717 | 3.9659 | 7.7555 | | 1.9433 | 3.9615 | 7.3571 | | 1.9724 | 3.9814 | 7.4137 |
| Efficiency (S / P) | | 0.9858 | 0.9914 | 0.9694 | | 0.9716 | 0.9903 | 0.9196 | | 0.9862 | 0.9953 | 0.9267 |

Next, we will plot the speedup and efficiency graphic in order to obtain the optimal number of nodes.

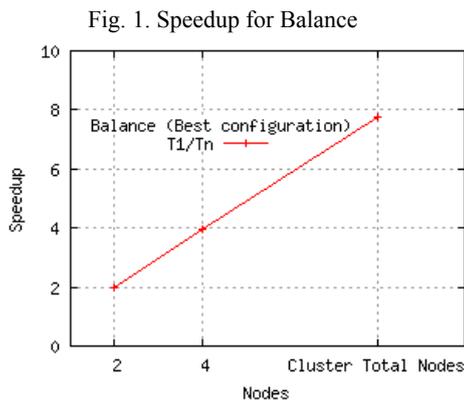

Fig. 1. Speedup for Balance

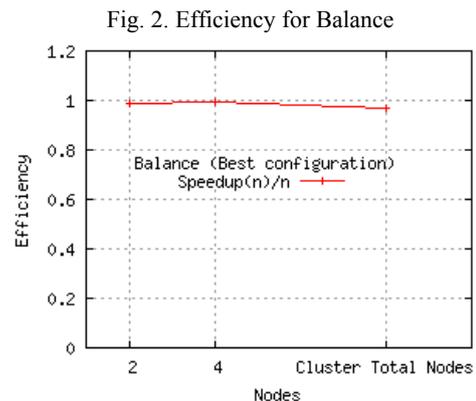

Fig. 2. Efficiency for Balance

As can be seen, the speedup does not increase linearly, but it saturates itself. In this case, the saturation takes place from 4 nodes on. We have plotted the efficiency values to establish a comparison with all values in the same scale. So, it is easier to obtain the optimal number of computing nodes; it is only necessary to verify at which number the efficiency value is greatest. The optimal number is 4, as was indicated in the speedup graph.

## 4. CONCLUSIONS

Once we have obtained the results of the different experiments we can conclude the following: a) two proposals have been implemented, that let distribute different features of the EA and the topology or the experimentation process; b) after the tests have been performed, we can mention that the efficiency and efficacy of the proposal are acceptable; c) we have carried out experiments with medium-size classification databases to show the model basic behaviour; d) by means of model distribution we have overcome the previous results.